\documentclass[runningheads]{llncs}
\usepackage[T1]{fontenc}
\usepackage{booktabs}
\usepackage{soul}
\usepackage{multirow}
\usepackage{arydshln}
\usepackage[export]{adjustbox}
\usepackage{graphicx}
\usepackage{enumitem}
\usepackage{amsmath}
\usepackage{amssymb}
\usepackage{bm}
\usepackage{algorithm}
\usepackage[noend]{algpseudocode}
\usepackage{caption}

\usepackage{xcolor}
\usepackage[normalem]{ulem}

\begin{document}

\title{VM-NeRF: Tackling Sparsity in NeRF with View Morphing}

\author{Matteo Bortolon\inst{1,2,3} \and Alessio Del Bue\inst{2} \and Fabio Poiesi\inst{1,2}}

\institute{
TeV, Fondazione Bruno Kessler \and
PAVIS, Istituto Italiano di Tecnologia \and
DISI, University of Trento \\
\email{\{mbortolon,poiesi\}@fbk.eu} \\
\email{\{alessio.delbue\}@iit.it}
}

\maketitle

\begin{abstract}
  NeRF aims to learn a continuous neural scene representation by using a finite set of input images taken from various viewpoints.
A well-known limitation of NeRF methods is their reliance on data: the fewer the viewpoints, the higher the likelihood of overfitting.
This paper addresses this issue by introducing a novel method to generate geometrically consistent image transitions between viewpoints using View Morphing. 
Our VM-NeRF approach requires no prior knowledge about the scene structure, as View Morphing is based on the fundamental principles of projective geometry.
VM-NeRF tightly integrates this geometric view generation process during the training procedure of standard NeRF approaches. 
Notably, our method significantly improves novel view synthesis, particularly when only a few views are available.
Experimental evaluation reveals consistent improvement over current methods that handle sparse viewpoints in NeRF models. 
We report an increase in PSNR of up to 1.8dB and 1.0dB when training uses eight and four views, respectively.
Source code: \url{https://github.com/mbortolon97/VM-NeRF}
\end{abstract}

%

\section{Introduction}\label{sec:introduction}

\begin{figure}[t]
\centering
\includegraphics[width=0.51\columnwidth]{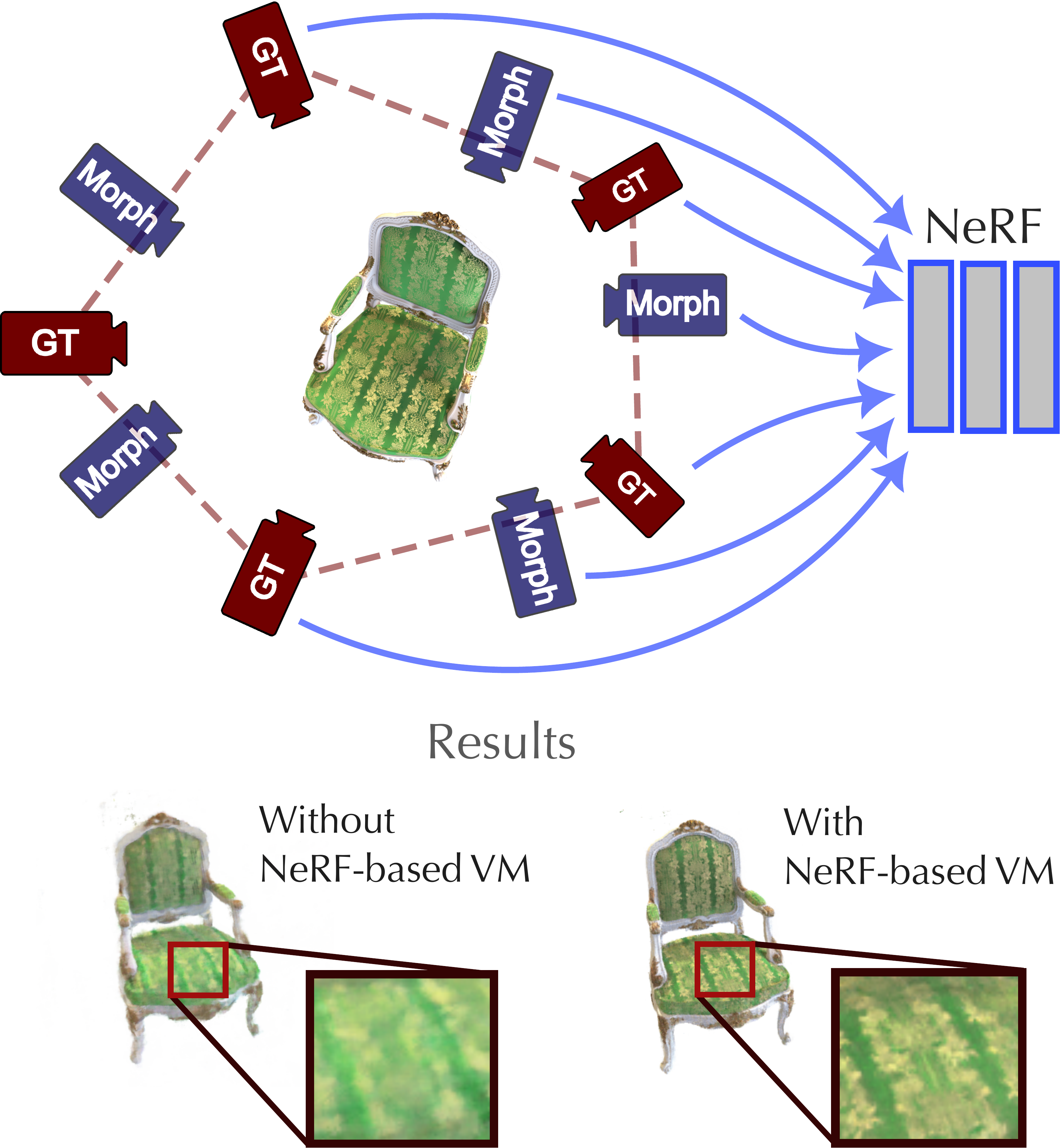}
\captionof{figure}{Given a set of known views (ground truth), View Morphing-NeRF (VM-NeRF) generates image transitions between views (morph) that can be effectively used to train a NeRF model in the case of few-shot view synthesis. Results are of a higher quality when VM-NeRF is used.}
\label{fig:teaser}
\end{figure}

Novel View Synthesis (NVS) is the problem of synthesising unseen camera views from a set of known views\footnote{Throughout the paper, we will use the term \textit{viewpoint} to refer to the camera pose, \textit{view} to refer to the scene seen through a certain viewpoint and to \textit{image} to refer to the photometric content captured from a view.} \cite{xie2022neural,orazio2020nvs}.
NVS is a key technology that can enable compelling augmented or virtual reality experiences~\cite{hedman2021snerg}, new entertainment technology~\cite{devernay2010novel}, and robotics applications~\cite{ichnowski2022dex}.
NVS has undergone a significant improvement after the introduction of Neural Radiance Fields (NeRF)~\cite{mildenhall2020nerf,barron2021mip} -- a trainable implicit neural representation of a 3D scene that can photorealistically render unseen (novel) views.
NeRF is a data-driven model that can synthesise high-quality novel views but in general requiring  several multi-view images, e.g.~about hundreds of images taken from different and uniformly distributed camera viewpoints around an object of interest \cite{mildenhall2020nerf}.
If these viewpoints are few and/or not uniformly distributed, the resulting NeRF model may fail to produce satisfactory novel views~\cite{jain2021dietnerf,mildenhall2019llff}.
This detrimental effect is a known drawback of NeRF-based approaches and it is due to the likelihood of overfitting on known viewpoints while decreasing generalisation on novel views that are furthest from the given viewpoints, namely the few-shot view synthesis problem~\cite{jain2021dietnerf}.



In this paper, we propose to tackle the problem of training a NeRF model on scenes captured with a sparse set of viewpoints by using a novel geometry-based strategy based on View Morphing~\cite{seitz1996viewMorphing} (Fig.~\ref{fig:teaser}). 
This purely geometric method can synthesise or morph a new viewpoint that lies in-between two given camera views while ensuring realistic image transitions. Traditionally, view morphing requires a set of accurate point matches between known image pairs in order to successfully perform the morph. As this matching stage is hard to integrate into a NeRF-based learning pipeline, our intuition is to leverage the per-image depth information implicitly estimated by NeRF to obtain dense coordinate matches among views after an image rectification stage (Fig.~\ref{fig:approach}). To this end, we have to relax and modify several steps of the view morphing strategy to be duly integrated in the NeRF learning paradigm. 
%
%
This technique does not require any prior knowledge about the captured 3D scene, and it can synthesise 3D projective transformations (e.g.~3D rotations, translations, shears) of objects by operating entirely on the input images.
We evaluate our approach by using the dataset of the original NeRF's paper \cite{mildenhall2020nerf} and we show that PSNR improves up to 1.8dB and 1.0dB when eight and four views are used for training, respectively.
We compare our approach with DietNeRF~\cite{jain2021dietnerf}, AugNeRF~\cite{chen2022augnerf} and RegNeRF~\cite{niemeyer2021regnerf}, and show that our approach can produce higher-quality renderings.

\noindent To summarise, our contributions are:
\setlist{nolistsep}
    \begin{itemize}[noitemsep]
    \item We present a novel and effective method for NeRF to address the problem of few-shot view synthesis;
    \item We introduce a new view morphing technique based on the NeRF depth output, named VM-NeRF;
    \item VM-NeRF can achieve higher-quality rendered images than alternative methods in the literature.
\end{itemize}
\section{Related work}\label{sec:related_work}

NVS scene synthesis can be solved either by using traditional 3D reconstruction techniques \cite{schoenberger2016sfm} or by adopting methods based on neural rendering \cite{Tewari2022}.
Neural Radiance Fields (NeRF) is a recent neural rendering method that can learn a volumetric representation of an unknown 3D scene approximating its radiance and density fields from a set of known (ground truth) views by using a multilayer perceptron (MLP) \cite{mildenhall2020nerf}.
NeRF optimises its parameters on one scene based on a set of known views, thus overfitting can occur when these views are few.

Current approaches addressing few-shot novel view synthesis can be divided into two groups.
The first group uses the same trained network to generate novel views of different scenes.
This category of methods trains on datasets characterised by similar scenes, such as DTU~\cite{aanaes2016dtumvs}.
Multiple-scene training can introduce datasets biases and may produce low-quality results in contexts outside the training domain~\cite{wang2022generalizingReview,muller2021realTimeNeuralCaching}.
SparseNeuS~\cite{long2022sparseneus} and ShaRF~\cite{rematas2021sharf} train NVS on multiple scenes by conditioning the MLP with features that encode appearance and geometry of the surface at a 3D location.
This can be achieved by using an auxiliary deep network jointly trained with NeRF.
The second group uses the original per-scene optimisation procedure of NeRF, so a single network trains and tests only on one scene leading to methods without dataset bias.
These methods are more likely to encounter overfit problems on the known views, however they reduce this likelihood by adding either semantic or geometric constraints during training.
DietNeRF belongs to this category and exploits the feature representations of known images computed with a CLIP pre-trained image encoder, renders random poses, and processes them by imposing semantic consistency through CLIP features~\cite{jain2021dietnerf}.
RegNeRF~\cite{niemeyer2021regnerf} renders random viewpoints around the known ones, and introduces regularisation constraints between known viewpoints and randomly sampled ones.

Single-scene methods working with few viewpoints may overfit on the known images, producing artefacts when novel views are rendered.
In general, we can mitigate overfitting via data augmentation~\cite{shorten2019dataAugmentation}, and to the best of our knowledge, the only methods that address data augmentation for NeRF are AugNeRF~\cite{chen2022augnerf} and GeoAug~\cite{chen2022geoaug}.
AugNeRF aims to improve NeRF generalisation by using adversarial data augmentation to enforce each ray and its augmented version to produce the same result.
GeoAug~\cite{chen2022geoaug} perturbs translation and rotation of the known viewpoints during training.
Our proposed approach does not perturb the known input views and rays, instead we create new views (novel 3D projective transformations) using pairs of known views.
This allows us to enforce coherence of newly rendered viewpoints between distant viewpoint pairs.
At the moment of the acceptance of this paper, we could not replicate the results of GeoAug because the authors have not released their source code.

\section{Preliminaries}

\subsection{NeRF overview}\label{sec:nerf_overview}

NeRF's objective is to synthesise novel views of a scene by optimising a volumetric function given a finite set of input views \cite{mildenhall2020nerf}.
Let \(f_{\bm{\theta}}\) be the underlying function we aim to optimise.
The input to \(f_{\bm{\theta}}\) is a 5D datum that encodes a point on a camera ray, i.e.~a 3D spatial location \((x,y,z)\) and a 2D viewing direction \((\theta, \phi)\).
Let \(\bm{c} \in \mathbb{R}^3\) be the view-dependent emitted radiance (colour) and \(\sigma\) be the volume density that \(f_{\bm{\theta}}\) predicts at \((x,y,z)\).
Novel views are synthesised by querying 5D data along the camera rays. 
Traditional volume rendering techniques can be used to transform \(\bm{c}\) and \(\sigma\) into an image \cite{Kajiya1984,Max1995}.
Because volume rendering is differentiable, \(f_{\bm{\theta}}\) can be implemented as a fully-connected deep network and learned.

Rendering a view from a novel viewpoint consists of estimating the integrals of all 3D rays that originate from the camera optic centre and that pass through each pixel of the camera image plane.
Let \(\bm{r}\) be a 3D ray.
To make rendering computationally tractable, each ray is represented as a finite set of 3D spatial locations, indexed with $i$, which are defined between two clipping distances: a near one ($t_n$) and a far one ($t_f$).
Let \(\Gamma\) be the number of 3D spatial locations sampled between \(t_n\) and \(t_f\).
Rendering the colour of a pixel is given by
\begin{equation}\label{eqn:rendering_formula}
\hat{\bm{c}}(\bm{r}) = \sum_{i=1}^{\Gamma} s(i) \left ( 1 - e^{-\hat{\bm{\sigma}}(\bm{r})_i \bm{\delta}_i} \right ) \hat{\bm{c}}(\bm{r})_i,
\end{equation}
where \(\hat{\bm{c}}(\bm{r})_i\) is the colour and \(\hat{\bm{\sigma}}(\bm{r})_i\) is the density predicted by the network at $i$.
$\bm{\delta}_i = t_{i+1} - t_i$ is the distance between adjacent sampled 3D spatial locations, and \(s(i)\) is the inverse of the volume density that is accumulated up to the \(i^{th}\) spatial location, which is in turn computed as
\begin{equation}\label{eqn:volume_density}
s(i) = e^{ - \sum^{i-1}_{j=1} \hat{\bm{\sigma}}(\bm{r})_j \bm{\delta}_j},
\end{equation}
where \((1 - e^{- \hat{\bm{\sigma}}(\bm{r})_i \bm{\delta}_i})\) is a density-based weight component: the higher the density value \(\sigma\) of a point, the larger the contribution on the final rendered colour.
 
Similarly to Eq.~\ref{eqn:rendering_formula}, we can render the pixel depth as
\begin{equation} \label{eqn:depth_nerf}
\bm{d}(\bm{r}) = \sum_{i=1}^{\Gamma} s(i) \left ( 1 - e^{- \hat{\bm{\sigma}}(\bm{r})_i \bm{\delta}_i} \right ) \bm{z}_i,
\end{equation}
where $\bm{z}_i$ is the distance of the $i^{th}$ spatial location with respect to the camera optic centre.

The input required to learn the NeRF parameters is a set of \(N\) images and their corresponding camera information.
Let \(\mathcal{I} = \{\bm{I}_k\}_{k=1}^N\) be the training images, and \(\mathcal{P} = \{\bm{P}_k\}_{k=1}^N\) and \(\mathcal{K} = \{\bm{K}_k\}_{k=1}^N\) be their corresponding camera poses and intrinsic parameters, respectively.
A pose $\bm{P} = [ \bm{R}, \bm{t} ]$ is composed of rotation $\bm{R}$ and translation $\bm{t}$.
We can estimate the depth map of a given view $k$ by rendering the depth of all its pixels, therefore we can define the estimated depth maps as $\mathcal{D} = \{\bm{D}_k\}_{k=1}^N$.

Learning \(f_{\bm{\theta}}\) is achieved by comparing each ground-truth pixel \(\bm{c}(\bm{r})\) with its predicted counterpart \(\hat{\bm{c}}(\bm{r})\).
The goal is to minimise the following L2-norm objective function
\begin{equation}\label{eqn:nerf_loss}
\mathcal{L} = \frac{1}{\lvert \mathcal{R} \rvert}\sum_{\bm{r} \in \mathcal{R}}
\left (
\lVert \bm{c}(\bm{r}) - \hat{\bm{c}}_c(\bm{r}) \rVert_2^2 +
\lVert \bm{c}(\bm{r}) - \hat{\bm{c}}_f(\bm{r}) \rVert_2^2
\right ),
\end{equation}
where \(\hat{\bm{c}}_c(\bm{r})\) and \(\hat{\bm{c}}_f(\bm{r})\) are the coarse and fine predicted volume colours for ray \(\bm{r}\), respectively.
Please refer to \cite{mildenhall2020nerf} for more details.

\subsection{View morphing overview}\label{sec:view_moprhing_overview}

View morphing objective is to synthesise natural 2D transitions between an image pair $\{\bm{I}_k, \bm{I}_{k'}\}$ and the approach can be summarised in three steps: 
\textit{i)} the two images are \textit{prewarped} through rectification, i.e.~their image planes are aligned without changing their cameras' optic centres; 
\textit{ii)} the \textit{morph} is computed between these prewarped images to generate a morphed image whose viewpoint lies on the line connecting the optic centres; 
\textit{iii)} the image plane of the morphed image is transformed to a desired viewpoint through \textit{postwarping}.

In practice, assuming the two views are prewarped, the morph uses the knowledge of their camera poses \(\bm{P}_k, \bm{P}_{k'}\), and the pixel correspondences between the images, i.e.~\(q_k: \bm{I}_k \Rightarrow \bm{I}_{k'}, q_{k'}: \bm{I}_{k'} \Rightarrow \bm{I}_k\) where \(q_k\) is a function that maps a pixel of \(\bm{I}_k\) to the corresponding pixel in \(\bm{I}_{k'}\)~\cite{seitz1996viewMorphing}.
Sparse pixel correspondences can be defined by a user or determined by a keypoint detector, they can then be densified via interpolation to create a dense correspondence map. 
\textit{This procedure is not viable as is in a learning-based pipeline, hence we have to define a novel view morphing strategy for a NeRF-based network architecture.}
A warp function for each image can be computed from the correspondence map through linear interpolation
\begin{align}\label{eqn:view_morphing_loss}
& \hat{\bm{I}}_{k,\alpha} = ( 1-\alpha)\hat{\bm{I}}_k + \alpha q_k ( \hat{\bm{I}}_k ) \nonumber\\
& \hat{\bm{I}}_{k',\alpha}  = ( 1 - \alpha ) q_{k'} ( \hat{\bm{I}}_{k'} ) + \alpha \hat{\bm{I}}_{k'} \nonumber\\
& \bm{P}_\alpha = (1 - \alpha) \bm{P}_0 + \alpha \bm{P}_1,
\end{align}
where $\hat{\bm{I}}_k$ are the coordinates of the image of camera $k$, 
$\hat{\bm{I}}_{k,\alpha}$ are pixel coordinates of the morphed image, 
and $\alpha \in [0,1]$ regulates the position of the morphed view along the line connecting the two views. 
The morphed image can then be computed by averaging the pixel colours of the warped images.
Please refer to \cite{seitz1996viewMorphing} for more details.

\section{NeRF-based View Morphing}

The goal of NeRF-based View Morphing (VM-NeRF) is to use the geometrical constraints of the morphing technique to synthesise a set of additional training input views $\mathcal{M} = \{ \textbf{M}_{(k,k'),\alpha}\}$, where $\textbf{M}_{(k,k'),\alpha}$ is a morphed view generated from the view pair $k$ and $k'$ with a given value of $\alpha$.
Adapting view morphing in a learning-based pipeline is challenging as we need reliable pixel correspondences ($q_k$ and $q_{k'}$) to synthesise morphed views.
Our intuition is that it is possible to compute one-to-one correspondences from the disparity information, a function of the depth as in Eq.~\ref{eqn:depth_nerf}, which we can render with the very same NeRF model.\sout{\color{red}!}
We can then linearly interpolate the photometric content of the view pair to produce the morphed view.

\begin{figure*}[t]
\centering
\includegraphics[width=1\textwidth]{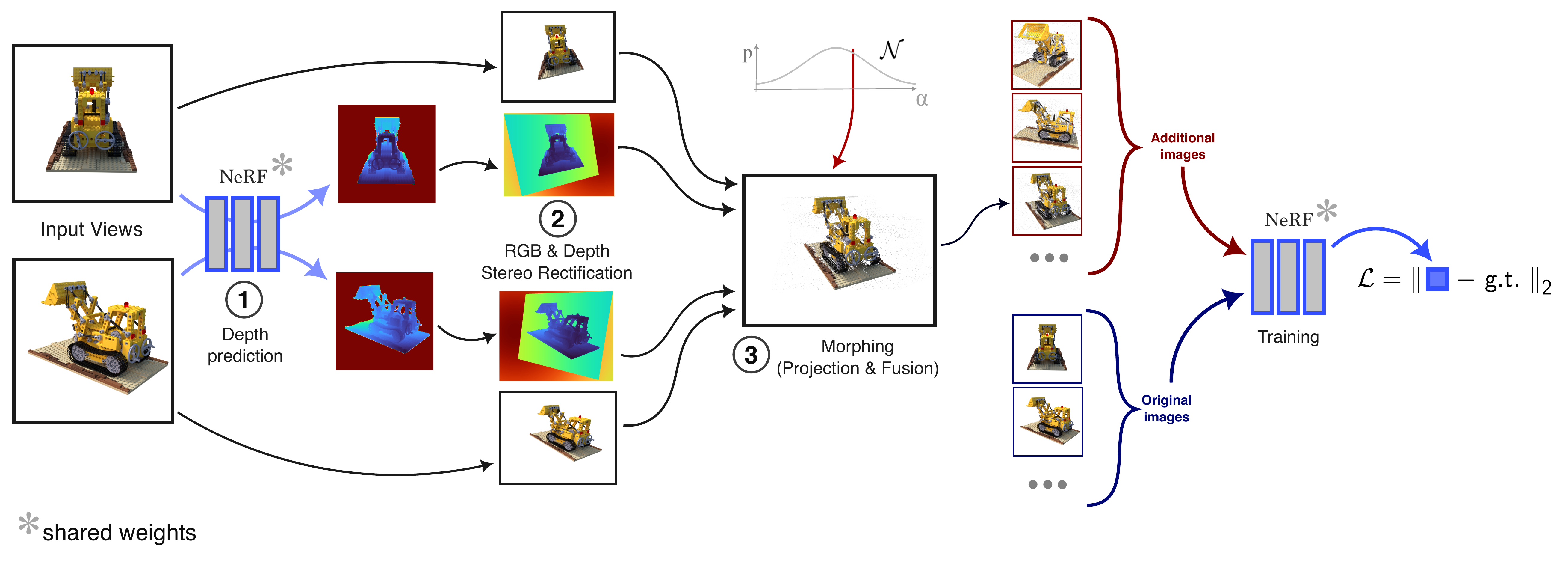}
\vspace{-.6cm}
\caption{Block diagram of NeRF-based View Morphing (VM-NeRF). 
From the left, we (1) predict the depth with NeRF, (2) rectify the input images and predicted depths, and (3) compute the image morphing of a view randomly positioned between the view pair.
\(\alpha\) determines the new view position and it is sampled from a Gaussian distribution.}
\label{fig:approach}
\end{figure*}

Based on the description in Sec.~\ref{sec:view_moprhing_overview}, we integrate in NeRF only the steps of prewarping and morphing.
We experimentally found that postwarping does not lead to better results.
Sec.~\ref{sec:rectification} describes how we perform the initial rectification of the two cameras.
Sec.~\ref{sec:image_morphing} describes how the images morphing is computed.
Sec.~\ref{sec:training_vm_nerf} provides detailed information on our practical approach to training NeRF with View Morphing.
Fig.~\ref{fig:approach} shows the block diagram of our approach.

\subsection{Rectification}\label{sec:rectification}

Our first step is rectification, which leads to rotating the known camera poses $\bm{P}_k$ and $\bm{P}_{k'}$ around their optic centres until their image planes become coplanar.
We can then compute the common image plane by using a selection of algorithms such as~\cite{fusiello2000stereoRectification,hartley2004multiview}.
We represent this plane as the rotation matrix
\begin{equation}\label{eqn:common_plane_equation}
    \tilde{\bm{R}} = [ \bm{a}_{x}, \bm{a}_{y}, \bm{a}_{z} ],
\end{equation}
where $\bm{a}_{x}, \bm{a}_{y}, \bm{a}_{z}$ are the axis components of the coplanar plane resulting from the rectification.
Stereo rectification is applied to the original images \( \left \{ \bm{I}_k, \bm{I}_{k'} \right \} \) and depth maps \( \left \{ \bm{D}_k, \bm{D}_{k'} \right \} \) predicted in Eq. \ref{eqn:depth_nerf}.
The new camera pose of view \(k\) is equal to $\tilde{\bm{P}}_k = [ \tilde{\bm{R}}, \bm{t}_k ]$, where $\bm{t}_k$ is the translation of the original camera pose $\bm{P}_k$ (same applies to view $k'$).

Rectification algorithms are typically based on the assumptions that viewpoints are aligned horizontally and that the reference viewpoint is the left-hand side of the camera (from an observer positioned behind the cameras)~\cite{fusiello2000stereoRectification,hartley2004multiview}.
This is atypical in NeRF, as viewpoints may have arbitrary camera configurations, leading to errors that should be corrected.
We mitigate this problem by comparing $\bm{a}_{z}$ with the $z$ component of the original view pose.
If this angle is greater than 45\textdegree\ with respect to both $\bm{P}_k$ and $\bm{P}_{k'}$, we rotate the warping matrices and poses by 90\textdegree\ or 180\textdegree.
The application of this modification to conventional rectification algorithm allows us to correctly generate the following rectified images $\{ \bm{\tilde{I}}_k, \bm{\tilde{I}}_{k'} \}$ and rectified depth maps $\{ \bm{\tilde{D}}_k, \bm{\tilde{D}}_{k'} \}$.

\subsection{Image morphing}\label{sec:image_morphing}

The second step is image morphing, i.e. fusing the rectified images to obtain the new morphed image. 
This procedure is divided in three steps: 
\textit{i)} finding the pixel correspondences; \textit{ii)} computing the position of each pixel on the morphed camera; 
\textit{iii)} fusing pixels that fall in the same position.
To determine the image correspondences, we initially compute the disparity maps as functions of the rectified estimated depths
\begin{equation}
\bm{E}_k = \frac{f_k}{\bm{\tilde{D}}_{k}} \left \| \bm{o}_k - \bm{o}_{k'} \right \|_{2},   
\quad       
\bm{E}_{k'} = \frac{f_{k'}} {\bm{\tilde{D}}_{k'}} \left \| \bm{o}_k - \bm{o}_{k'} \right \|_{2},
\end{equation}
where $\{ \bm{o}_k, \bm{o}_{k'} \}$ are the principal points and $\{f_k, f_{k'} \}$ are the focal lengths of cameras $k$ and $k'$.

Then, we determine the correspondences of the pixel positions between images defined in Eq.~\ref{eqn:view_morphing_loss} as
\begin{equation}
q_k(\hat{\bm{I}}_k) = \hat{\bm{I}}_k + \frac{\bm{\tilde{b}}_k}{\| \bm{\tilde{b}}_k \|_2} \mathbf{1}^\top \odot \bm{E}_k,
\end{equation}
where $\mathbf{1}$ is a vector of ones, $\odot$ indicates the Hadamard product and $\hat{\bm{I}}_k$ is the baseline direction with respect to the common plane defined in Eq.~\ref{eqn:common_plane_equation} that is computed as
\begin{equation}
\bm{\tilde{b}}_k = \bm{a}_z \times ( (\bm{o}_k - \bm{o}_{k'}) \times \bm{a}_z ).
\end{equation}
The same operation is computed for $k'$.
Then, we apply the warp functions of Eq.~\ref{eqn:view_morphing_loss} to compute the position of each pixel on the morphed view, thus obtaining $\hat{\bm{I}}_{k,\alpha}$ and $\hat{\bm{I}}_{k',\alpha}$.

Lastly, a coalescence operation~\cite{chaitin1982registerCoalescence} fuses the pixels of the two views $k$ and $k'$.
The coalescence operation concatenates two sets of coordinates and fuse pixels with the same position, preserving only the pixel values of the points that are nearest to the camera.
We use $\{ \tilde{\bm{D}}_k, \tilde{\bm{D}}_{k'} \}$ to determine the distance of the points.

\subsection{Training with VM-NeRF}\label{sec:training_vm_nerf}

VM-NeRF is subject to the same geometric constraints as the original view morphing technique \cite{seitz1996viewMorphing}.
These constraints impose that singular camera configurations should not exist. These configurations happen whenever the optic centre of a camera is within the field of view of another one \cite{seitz1996viewMorphing}.
We also discard cameras that are distant from each other more than a threshold \(\gamma\), as the morphed cameras may be on a transition path that crosses regions where the object of interest is not actually visible (so being rather useless for training a NeRF based model).

Because view morphing allows the synthesis of a new view at any point on the line that connects the known camera pair, we randomly sample new views using a Gaussian distribution centred halfway through the camera pair.
Specifically, let us consider a normalised distance between the two cameras. The Gaussian distribution is centred at 0.5 and the standard deviation \(\sigma\) is chosen such that \(3\sigma \rightarrow \epsilon\) at the optic centre positions.
Therefore, we sample \(\alpha \sim \mathcal{N}(0.5, \sigma)\) with $0 \le \alpha \le 1$.
The depth NeRF can render at the first few iterations is noisy, therefore, we let NeRF warm up on the known views for \(\lambda\) iterations before synthesising and injecting VM-NeRF  views in the next training iterations.
After the warm-up, for each valid camera pair, we regenerate \(M\) new views every \(\eta\) training iterations as the predicted depth improves over time during training.

\section{Experiments}

\begin{figure}[t]
  \begin{minipage}{0.45\linewidth}
\caption{Results on the NeRF realistic synthetic 360$^\circ$ dataset.}
\label{tab:overview_table}
    \centering
    \resizebox{.92\columnwidth}{!}{
    \begin{tabular}{llrrr}
    \toprule
    \# views &  Method & PSNR $\uparrow$ & SSIM $\uparrow$ & LPIPS $\downarrow$ \\
    \midrule
    100 \cite{mildenhall2020nerf} & NeRF \cite{mildenhall2020nerf} & 31.21 & 0.9513 & 0.0465 \\
    \midrule
    \multirow{5}{*}{8} & NeRF \cite{mildenhall2020nerf} & 23.45 & 0.8673 & 0.1303 \\
     & DietNeRF \cite{jain2021dietnerf} & 22.98 & 0.8545 & 0.1258 \\
     & AugNeRF \cite{chen2022augnerf} & 10.04 & 0.5415 & 0.3866 \\
     & RegNeRF \cite{niemeyer2021regnerf} & 22.91 & 0.8756 & 0.1138 \\
     & VM-NeRF & \textbf{24.39} & \textbf{0.8768} & \textbf{0.1146} \\
    \midrule
    \multirow{3}{*}{8 \cite{jain2021dietnerf}} & NeRF \cite{jain2021dietnerf} & 20.09 & 0.8220 & 0.1790 \\
    & DietNeRF \cite{jain2021dietnerf} & 23.59 & 0.8740 & 0.0970 \\
    & VM-NeRF & \textbf{24.14} & \textbf{0.8729} & \textbf{0.1180} \\
    \midrule
    \multirow{5}{*}{4} & NeRF \cite{mildenhall2020nerf} & 10.98 & 0.6550 & 0.3620 \\
     & DietNeRF \cite{jain2021dietnerf} & 12.61 & 0.6591 & 0.3302 \\
     & AugNeRF \cite{chen2022augnerf} & 8.14 & 0.3924 & 0.4802 \\
     & RegNeRF \cite{niemeyer2021regnerf} & 15.88 & \textbf{0.7932} & \textbf{0.1994} \\
     & VM-NeRF & \textbf{16.90} & 0.7563 & 0.2461 \\
    \bottomrule
    \end{tabular}
    }
  \end{minipage}\hfill
  \begin{minipage}{0.45\linewidth}
\vspace{-4.9em}
\caption{Ablation study results. Keys: \# views: represent the number of views and the relative subset. avg. dist.: average distance between view pairs.}
\label{tab:chair_scene_dvm_behaviour}
\resizebox{\columnwidth}{!}{
\begin{tabular}{lcllrr}
\toprule
\# views & avg. dist. & Method & PSNR $\uparrow$ & SSIM $\uparrow$ & LPIPS $\downarrow$ \\ 
\midrule
8 \cite{jain2021dietnerf} & 5.20 & DietNeRF \cite{jain2021dietnerf} & 25.59 & 0.9120 & 0.0770 \\
8 \cite{jain2021dietnerf} & 5.20 & VM-NeRF & 26.90 & 0.9180 & 0.0797 \\
8 (s1) & 5.18 & VM-NeRF & 27.87 & 0.9294 & 0.0693 \\
8 (s2) & 4.38 & VM-NeRF & 27.13 & 0.9178 & 0.0855 \\
8 (s3) & 4.73 & VM-NeRF & 27.48 & 0.9317 & 0.0676 \\
8 (s4) & 4.69 & VM-NeRF & 28.39 & 0.9365 & 0.0598 \\
\bottomrule
\end{tabular}
}

  \end{minipage}
\end{figure}

\subsection{Experimental setup}

We evaluate our method on three training setups  using the NeRF realistic synthetic 360\textdegree dataset \cite{mildenhall2020nerf}, which is composed of eight scenes, i.e.~Chair, Drums, Ficus, Lego, Materials, Ship, Mic, Hot Dog.
\textbf{First setup:} We select \(N=8\) views out of 100 available for each scene using the Farthest Point Sampling (FPS) \cite{qi2019deepHoughVoting} (the first view is used for FPS initialisation in each scene).
\textbf{Second setup:} we use the same \(N=8\) views used in DietNeRF \cite{jain2021dietnerf}.
\textbf{Third setup:} we select \(N=4\) views using the previous FPS approach.
We test each trained model on all the test views of NeRF realistic synthetic 360\textdegree.
We quantify the rendering results using the peak signal-to-noise ratio (PSNR) score, the structured similarity index measure (SSIM) \cite{wang2004ssim} and the learned perceptual image patch similarity (LPIPS) \cite{zhang2018perceptual}.
We quantitatively compare our approach against DietNeRF~\cite{jain2021dietnerf} and RegNeRF~\cite{niemeyer2021regnerf} as the most recent methods for few-shot view synthesis. 
We also compare against AugNeRF \cite{chen2022augnerf} because it is the only data augmentation for NeRF, and data augmentation can be a useful strategy to promote generalisation.  
We choose to use the Chair scene for our ablation study, which consists of testing VM-NeRF on four different, randomly-chosen, configurations of eight views and on the DietNeRF configuration.

We implement NeRF and our approach in PyTorch Lightning, and run experiments on a single Nvidia A40 with a batch size of 1024 rays. A single scene can be trained in about two days. We use the original implementations of DietNeRF, AugNeRF and RegNeRF to evaluate the different setups. 
We set the same training parameters as in \cite{mildenhall2020nerf}, and set \(\gamma = 6\), \(\sigma = 0.2\), \(M=1\), \(\eta = 5\), \(\lambda = 500\).

\subsection{Analysis of the results}

\noindent \textbf{Quantitative.} 
Tab.~\ref{tab:overview_table} shows the results averaged over the eight scenes.
Our NeRF implementation can achieve nearly the same results reported in \cite{mildenhall2020nerf} on the 100-view setup, 
i.e.~PSNR equal to 31.21 (ours) compared to 31.01 \cite{mildenhall2020nerf}.

VM-NeRF can outperform all the other methods in the eight-view setting.
Interestingly, the original version of NeRF is the one that performs as second best, followed by DietNeRF and RegNeRF.
AugNeRF fails to produce satisfactorily results.
We can also observe that VM-NeRF achieves slightly better quality than its version with oracle depth maps, i.e.~24.39 vs.~24.22 PSNR.
In fact, we observed that VM-NeRF can effectively leverage the depth information that is estimated during training, although it is noisy.
We also evaluate VM-NeRF on the same eight views originally tested by DietNeRF \cite{jain2021dietnerf}.
Also here we can achieve higher quality results on average, i.e.~24.14 vs.~23.59.
We also improve in the four-view setup where we obtain an improvement of +1.02 PSNR on average.
The results also show that the perturbation of the known input views, done by AugNeRF, has adverse effects in all the tested setups.

\sethlcolor{white}
\begin{figure*}[!t]
    \centering
    \resizebox{0.9\textwidth}{!}{
    \begin{tabular}{ccccccc}
         & \footnotesize{Ground Truth} & \footnotesize{NeRF} \cite{mildenhall2020nerf} & \footnotesize{AugNeRF} \cite{chen2022augnerf} & \footnotesize{DietNeRF} \cite{jain2021dietnerf} & \footnotesize{RegNeRF} \cite{niemeyer2021regnerf} & \footnotesize{VM-NeRF} \\\noalign{\medskip}
        \multirow{3}{*}{\small{\begin{tabular}{@{}c@{}}Chair \\ \includegraphics[width=.2\linewidth,valign=m]{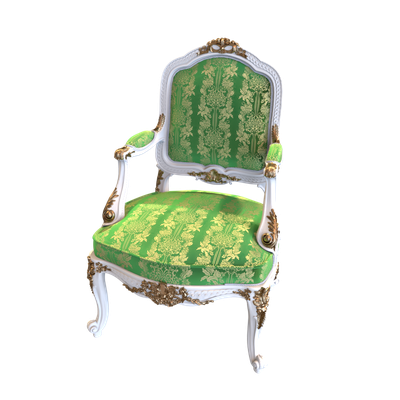} \end{tabular}}} &  & \footnotesize{23.52} & \footnotesize{11.94} & \footnotesize{22.74} & \footnotesize{20.91} & \footnotesize{25.80} \\
        & \includegraphics[width=.15\linewidth,valign=m,trim={200px 169px 76px 107px},clip]{images/raw_images/gt_chair_55.png} & \includegraphics[width=.15\textwidth,valign=m,trim={200px 169px 76px 107px},clip]{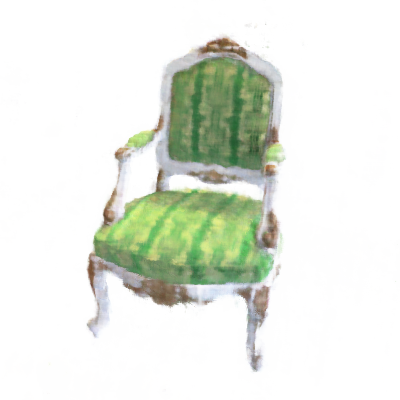} & \includegraphics[width=.15\linewidth,valign=m,trim={200px 169px 76px 107px},clip]{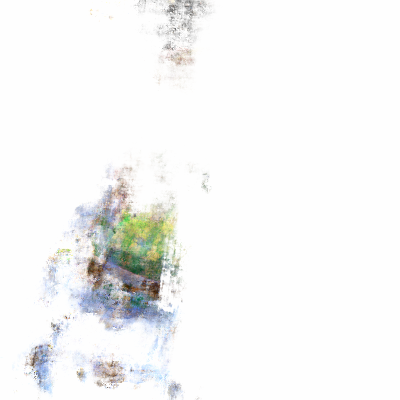} & \includegraphics[width=.15\linewidth,valign=m,trim={200px 169px 76px 107px},clip]{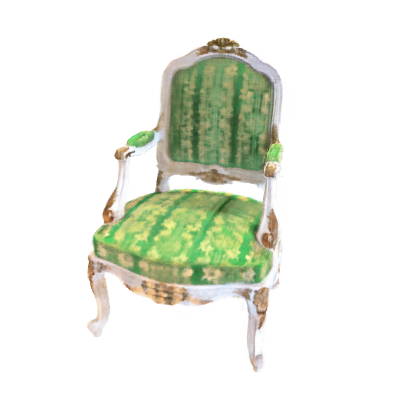} & \includegraphics[width=.15\linewidth,valign=m,trim={200px 169px 76px 107px},clip]{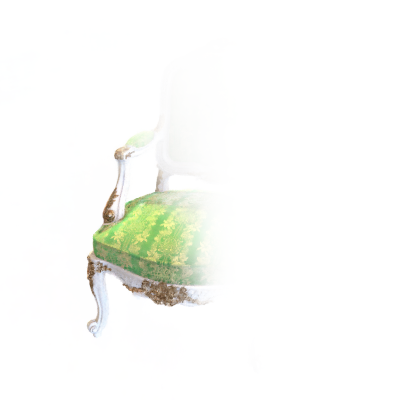} & \includegraphics[width=.15\linewidth,valign=m,trim={200px 169px 76px 107px},clip]{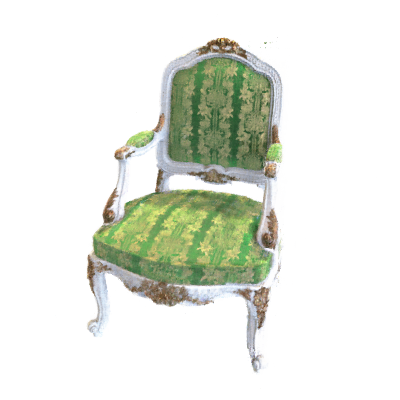} \\ \noalign{\medskip}
        & \includegraphics[width=.15\linewidth,valign=m,trim={127px 84px 128px 171px},clip]{images/raw_images/gt_chair_55.png} & \includegraphics[width=.15\linewidth,valign=m,trim={127px 84px 128px 171px},clip]{images/result_images/nerf_chair_55.png} &  \includegraphics[width=.15\linewidth,valign=m,trim={127px 84px 128px 171px},clip]{images/result_images/augnerf_chair_55.png} & \includegraphics[width=.15\linewidth,valign=m,trim={127px 84px 128px 171px},clip]{images/result_images/dietnerf_chair_55.png} & \includegraphics[width=.15\linewidth,valign=m,trim={127px 84px 128px 171px},clip]{images/result_images/regnerf_chair_55.png} & \includegraphics[width=.15\linewidth,valign=m,trim={127px 84px 128px 171px},clip]{images/result_images/dvm_chair_55.png} \\ \noalign{\medskip}
        \multicolumn{1}{c}{\multirow{3}{*}{\small{\begin{tabular}{@{}c@{}}Hot Dog \\ \includegraphics[width=.2\linewidth,valign=m]{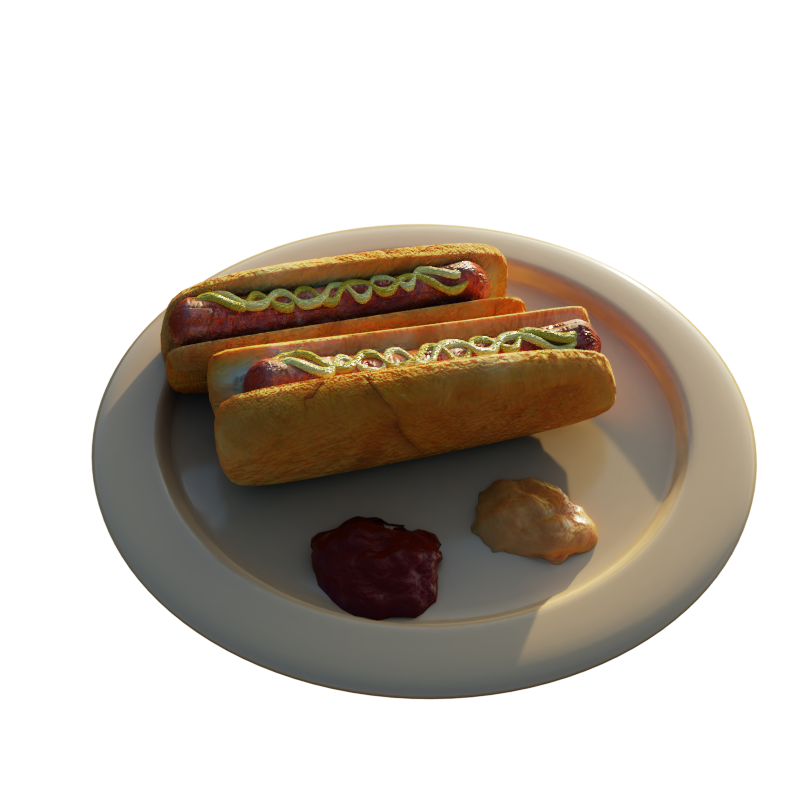} \end{tabular}}}} & & \footnotesize{27.59} & \footnotesize{12.45} & \footnotesize{28.76} & \footnotesize{30.20} & \footnotesize{31.29} \\ & \includegraphics[width=.15\linewidth,valign=m,trim={300px 84px 188px 404px},clip]{images/raw_images/gt_hotdog_21.png} & \includegraphics[width=.15\linewidth,valign=m,trim={150px 42px 94px 202px},clip]{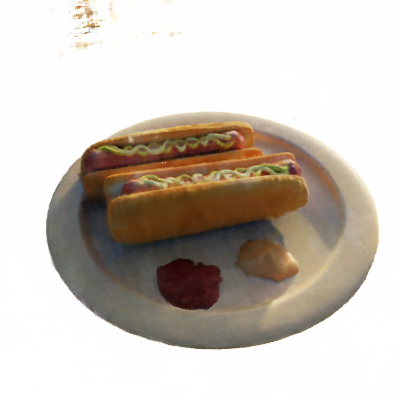} & \includegraphics[width=.15\linewidth,valign=m,trim={150px 42px 94px 202px},clip]{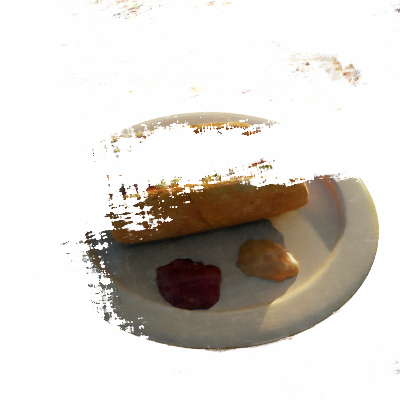} & \includegraphics[width=.15\linewidth,valign=m,trim={150px 42px 94px 202px},clip]{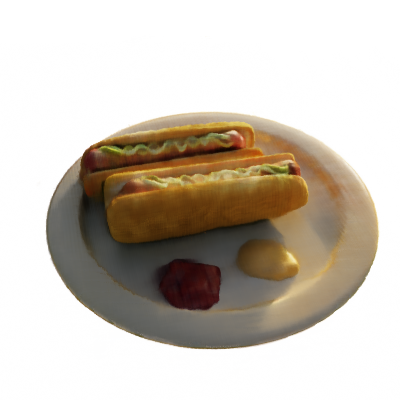} & \includegraphics[width=.15\linewidth,valign=m,trim={150px 42px 94px 202px},clip]{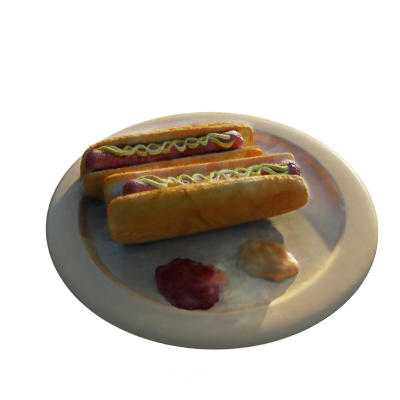} & \includegraphics[width=.15\linewidth,valign=m,trim={150px 42px 94px 202px},clip]{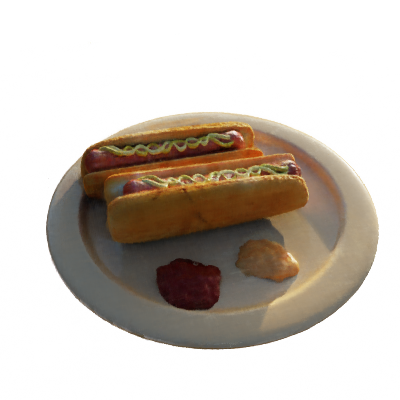} \\\noalign{\medskip}
        & \includegraphics[width=.15\linewidth,valign=m,trim={400px 374px 196px 222px},clip]{images/raw_images/gt_hotdog_21.png} &  \includegraphics[width=.15\linewidth,valign=m,trim={200px 187px 98px 111px},clip]{images/result_images/nerf_hotdog_21.png} & \includegraphics[width=.15\linewidth,valign=m,trim={200px 187px 98px 111px},clip]{images/result_images/augnerf_hotdog_21.png} & \includegraphics[width=.15\linewidth,valign=m,trim={200px 187px 98px 111px},clip]{images/result_images/dietnerf_hotdog_21.png} & \includegraphics[width=.15\linewidth,valign=m,trim={200px 187px 98px 111px},clip]{images/result_images/regnerf_hotdog_21.png} & \includegraphics[width=.15\linewidth,valign=m,trim={200px 187px 98px 111px},clip]{images/result_images/dvm_hotdog_21.png} \\\noalign{\medskip}
        \multirow{2}{*}{\small{\begin{tabular}{@{}c@{}} Lego \\ \includegraphics[width=.2\linewidth,valign=m]{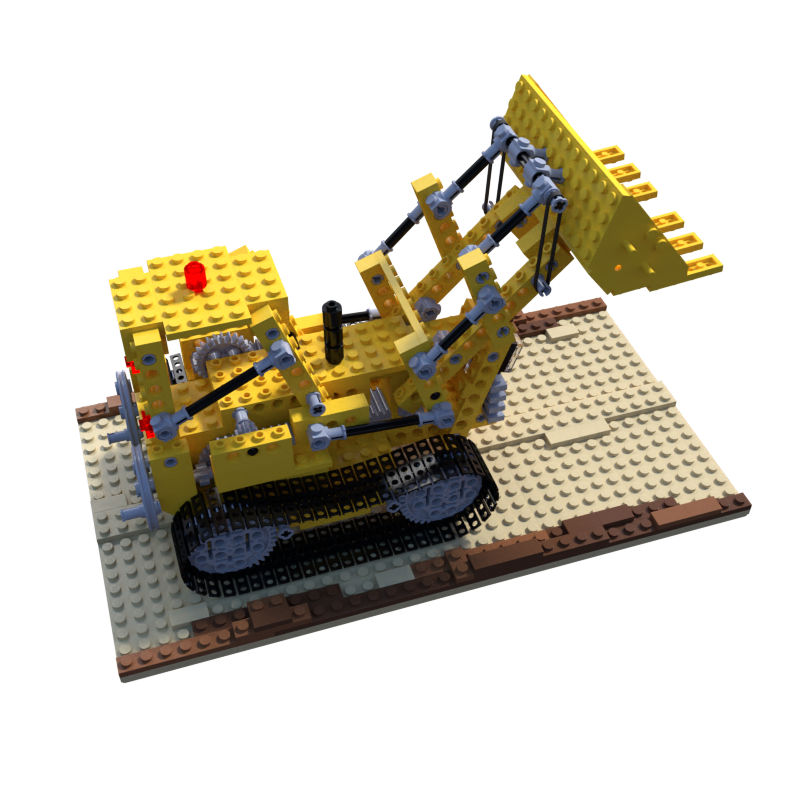} \end{tabular}}} &  & \footnotesize{22.12} & \footnotesize{1.38} & \footnotesize{18.61} & \footnotesize{22.85} & \footnotesize{22.20} \\ & \includegraphics[width=.15\linewidth,valign=m,trim={100px 396px 496px 200px},clip]{images/raw_images/gt_lego_20.png} &  \includegraphics[width=.15\linewidth,valign=m,trim={50px 198px 248px 100px},clip]{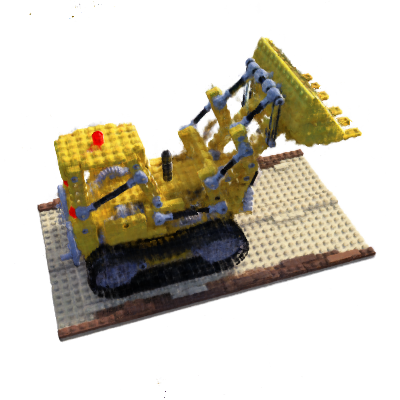} & \includegraphics[width=.15\linewidth,valign=m,trim={50px 198px 248px 100px},clip]{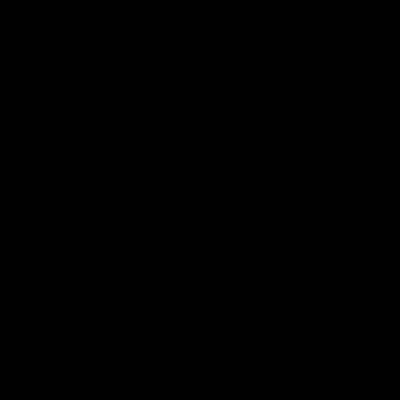} & \includegraphics[width=.15\linewidth,valign=m,trim={50px 198px 248px 100px},clip]{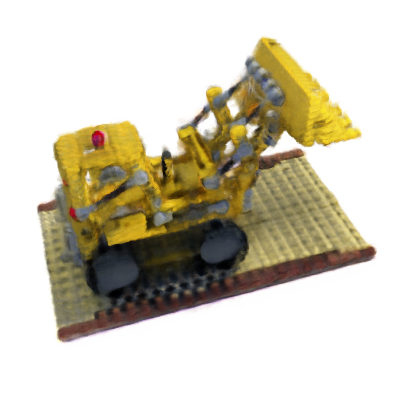} & \includegraphics[width=.15\linewidth,valign=m,trim={50px 198px 248px 100px},clip]{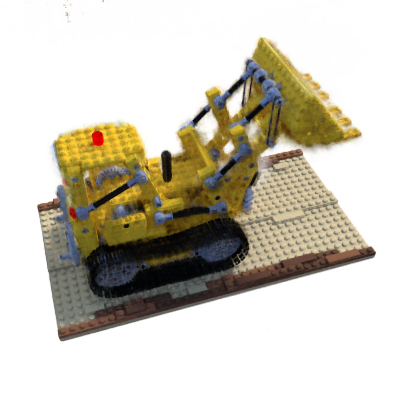} & \includegraphics[width=.15\linewidth,valign=m,trim={50px 198px 248px 100px},clip]{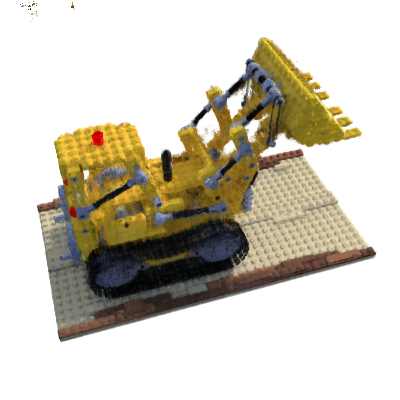} 
        \\\noalign{\medskip}
        & \includegraphics[width=.15\linewidth,valign=m,trim={148px 154px 448px 442px},clip]{images/raw_images/gt_lego_20.png} &  \includegraphics[width=.15\linewidth,valign=m,trim={74px 77px 224px 221px},clip]{images/result_images/nerf_lego_20.png} & \includegraphics[width=.15\linewidth,valign=m,trim={74px 77px 224px 221px},clip]{images/result_images/augnerf_lego_20.png} & \includegraphics[width=.15\linewidth,valign=m,trim={74px 77px 224px 221px},clip]{images/result_images/dietnerf_lego_20.png} & \includegraphics[width=.15\linewidth,valign=m,trim={74px 77px 224px 221px},clip]{images/result_images/regnerf_lego_20.png} & \includegraphics[width=.15\linewidth,valign=m,trim={74px 77px 224px 221px},clip]{images/result_images/dvm_lego_20.png} \\ \noalign{\medskip}

    \end{tabular}
    \setlength{\tabcolsep}{6pt}
    }
    \caption{
    Comparisons on test-set views of scenes of NeRF realistic synthetic 360\textdegree. 
    Unlike AugNeRF~\cite{chen2022augnerf}, VM-NeRF is an effective method that can be used for few-shot view synthesis problems.
    Unlike DietNeRF~\cite{jain2021dietnerf}, VM-NeRF enables NeRF to learn scenes with a higher definition.
    VM-NeRF produce less artefacts than RegNeRF during rendering~\cite{niemeyer2021regnerf}.
    We report the PSNR that we measured for each method and for each rendered image.
    AugNeRF unsuccessfully learns Chair and Lego (white and black outputs).}
    \label{fig:8_views_qualitative_results}
\end{figure*}

\noindent \textbf{Qualitative.} 
Fig.~\ref{fig:8_views_qualitative_results} shows some qualitative results on Chair, Hot Dog and Lego where we can observe that VM-NeRF produce results with better details than DietNeRF.
We speculate that this difference with DietNeRF may be due to its CLIP-based approach that is introduced to leverage a semantic consistency loss for regularisation~\cite{Radford2021clip}. The CLIP output is a low-dimensional (global) representation vector of the image, which may hinder the learning of high-definition details.
Differently, our approach interpolates the original photometric information from two views to produce a new input view, without losing information through the encoding of the low-dimensional representation vector.
Fig.~\ref{fig:8_views_qualitative_results} shows that our approach  compared to RegNeRF produces fewer artefacts by correlating the nearby views.

\noindent \textbf{Ablation study}
We assess the stability of VM-NeRF by evaluating the rendering quality when different combinations of views are used to train NeRF.
Tab.~\ref{tab:chair_scene_dvm_behaviour} shows that the performance is fairly stable throughout different view configurations.
We also observed that the algorithm is robust to variations in the distance between view pairs.
As long as a view pair is not singular and the distance between cameras is adequate to create acceptable 3D projective transformations of the object, we can successfully synthesise new views with VM-NeRF.

\section{Conclusions}\label{sec:conclusions}

We presented a novel method for few-shot view synthesis that blends NeRF and the View Morphing technique~\cite{seitz1996viewMorphing}.
View morphing requires no prior knowledge of the 3D shape and it is based on general principles of projective geometry.
We evaluated our approach using the conventional dataset employed by NeRF-based methods, demonstrating that VM-NeRF more effectively learns 3D scenes across various few-shot view synthesis setups.
VM-NeRF can interpolate only along the line that connects the optical centres of each camera pair.
Therefore, it cannot reconstruct the whole object if only a part of it is viewed during training.
Lastly, we designed our approach to be fully differentiable, so an attractive research direction is to integrate our approach into an end-to-end training pipeline.

\section*{Acknowledgements}\label{sec:acknowledgements}
This work was supported by the PNRR project FAIR - Future AI Research (PE00000013), under the NRRP MUR program funded by the NextGenerationEU.
This research is partially supported by the project Future Artificial Intelligence Research (FAIR) – PNRR MUR Cod. PE0000013 - CUP: E63C2200194 0006 and the framework "RAISE - Robotics and AI for Socio-economic Empowerment” supported by European Union - NextGenerationEU.

\bibliographystyle{splncs04}
\bibliography{egbib}

\begin{thebibliography}{10}
\providecommand{\url}[1]{\texttt{#1}}
\providecommand{\urlprefix}{URL }
\providecommand{\doi}[1]{https://doi.org/#1}

\bibitem{aanaes2016dtumvs}
Aan{\ae}s, H., Jensen, R.R., Vogiatzis, G., Tola, E., Dahl, A.B.: {Large-Scale
  Data for Multiple-View Stereopsis}. International Journal of Computer Vision
  \textbf{120}(2),  153--168 (Nov 2016)

\bibitem{barron2021mip}
Barron, J.T., Mildenhall, B., Tancik, M., Hedman, P., Martin-Brualla, R.,
  Srinivasan, P.P.: {Mip-NeRF: A multiscale representation for anti-aliasing
  neural radiance fields}. In: ICCV (2021)

\bibitem{chaitin1982registerCoalescence}
Chaitin, G.J.: {Register allocation \& spilling via graph coloring}. ACM
  Sigplan Notices  \textbf{17}(6),  98--101 (Jun 1982)

\bibitem{chen2022geoaug}
Chen, D., Liu, Y., Huang, L., Wang, B., Pan, P.: {GeoAug: Data Augmentation for
  Few-Shot NeRF with Geometry Constraints}. In: ECCV (2022)

\bibitem{chen2022augnerf}
Chen, T., Wang, P., Fan, Z., Wang, Z.: {Aug-NeRF: Training Stronger Neural
  Radiance Fields with Triple-Level Physically-Grounded Augmentations}. In:
  CVPR (2022)

\bibitem{devernay2010novel}
Devernay, F., Peon, A.R.: {Novel view synthesis for stereoscopic cinema:
  detecting and removing artifacts}. In: Workshop on 3D Video Processing
  (ACMMM) (2010)

\bibitem{fusiello2000stereoRectification}
Fusiello, A., Trucco, E., Verri, A.: {A compact algorithm for rectification of
  stereo pairs}. Machine Vision and Applications  \textbf{12}(1),  16--22 (Jul
  2000)

\bibitem{orazio2020nvs}
Gallo, O., Troccoli, A., Jampani, V.: {Novel View Synthesis: From Depth-Based
  Warping to Multi-Plane Images and Beyond} (2020),
  \url{https://nvlabs.github.io/nvs-tutorial-cvpr2020/}, conference on Computer
  Vision and Pattern Recognition

\bibitem{hartley2004multiview}
Hartley, R.I., Zisserman, A.: {Multiple View Geometry in Computer Vision}.
  Cambridge University Press (2004)

\bibitem{hedman2021snerg}
Hedman, P., Srinivasan, P.P., Mildenhall, B., Barron, J.T., Debevec, P.:
  {Baking Neural Radiance Fields for Real-Time View Synthesis}. In: ICCV (2021)

\bibitem{ichnowski2022dex}
Ichnowski, J., Avigal, Y., Kerr, J., Goldberg, K.: {Dex-NeRF: Using a Neural
  Radiance Field to Grasp Transparent Objects}. In: CRL (2022)

\bibitem{jain2021dietnerf}
Jain, A., Tancik, M., Abbeel, P.: {Putting NeRF on a Diet: Semantically
  Consistent Few-Shot View Synthesis}. In: ICCV (2021)

\bibitem{Kajiya1984}
Kajiya, J., Herzen, B.: {Ray tracing volume densities}. In: SIGGRAPH (1984)

\bibitem{long2022sparseneus}
Long, X., Lin, C., Wang, P., Komura, T., Wang, W.: {SparseNeuS: Fast
  Generalizable Neural Surface Reconstruction from Sparse views}. In: ECCV
  (2022)

\bibitem{Max1995}
Max, N.: {Optical models for direct volume rendering}. IEEE Transaction on
  Visualization and Computer Graphics  \textbf{1}(2),  99--108 (Jun 1995)

\bibitem{mildenhall2019llff}
Mildenhall, B., Srinivasan, P.P., Ortiz-Cayon, R., Kalantari, N.K.,
  Ramamoorthi, R., Ng, R., Kar, A.: {Local Light Field Fusion: Practical View
  Synthesis with Prescriptive Sampling Guidelines}. ACM Transactions on
  Graphics  \textbf{38}(29),  1--14 (Aug 2019)

\bibitem{mildenhall2020nerf}
Mildenhall, B., Srinivasan, P.P., Tancik, M., Barron, J.T., Ramamoorthi, R.,
  Ng, R.: {NeRF: Representing Scenes as Neural Radiance Fields for View
  Synthesis}. In: ECCV (2020)

\bibitem{muller2021realTimeNeuralCaching}
M{\"u}ller, T., Rousselle, F., Nov{\'a}k, J., Keller, A.: {Real-time neural
  radiance caching for path tracing}. ACM Transactions on Graphics
  \textbf{40}(4),  1--16 (Aug 2021)

\bibitem{niemeyer2021regnerf}
Niemeyer, M., Barron, J.T., Mildenhall, B., Sajjadi, M.S.M., Geiger, A.,
  Radwan, N.: {RegNeRF: Regularizing Neural Radiance Fields for View Synthesis
  from Sparse Inputs}. In: CVPR (2022)

\bibitem{qi2019deepHoughVoting}
Qi, C.R., Litany, O., He, K., Guibas, L.J.: {Deep hough voting for 3d object
  detection in point clouds}. In: ICCV (2019)

\bibitem{Radford2021clip}
Radford, A., Kim, J.W., Hallacy, C., Ramesh, A., Goh, G., Agarwal, S., Sastry,
  G., Askell, A., Mishkin, P., Clark, J., Krueger, G., Sutskever, I.: {Learning
  Transferable Visual Models From Natural Language Supervision}. In: ICML
  (2021)

\bibitem{rematas2021sharf}
Rematas, K., Martin-Brualla, R., Ferrari, V.: {ShaRF: Shape-conditioned
  Radiance Fields from a Single View}. In: ICML (2021)

\bibitem{schoenberger2016sfm}
Sch\"{o}nberger, J.L., Frahm, J.M.: {Structure-from-Motion Revisited}. In: CVPR
  (2016)

\bibitem{seitz1996viewMorphing}
Seitz, S.M., Dyer, C.R.: {View morphing}. In: Conference on Computer graphics
  and interactive techniques (1996)

\bibitem{shorten2019dataAugmentation}
Shorten, C., Khoshgoftaar, T.M.: {A survey on Image Data Augmentation for Deep
  Learning}. Journal of Big Data  \textbf{6}(1) (Jul 2019)

\bibitem{Tewari2022}
Tewari, A., et~al.: {Advances in Neural Rendering}. Computer Graphics Forum
  \textbf{41}(2),  703--735 (May 2022)

\bibitem{wang2022generalizingReview}
Wang, J., Lan, C., Liu, C., Ouyang, Y., Qin, T., Lu, W., Chen, Y., Zeng, W.,
  Yu, P.S.: {Generalizing to Unseen Domains: A Survey on Domain
  Generalization}. IEEE Transactions on Knowledge \& Data Engineering
  \textbf{35}(08),  8052--8072 (Aug 2023)

\bibitem{wang2004ssim}
Wang, Z., Bovik, A., Sheikh, H., Simoncelli, E.: {Image quality assessment:
  from error visibility to structural similarity}. IEEE Transactions on Image
  Processing  \textbf{13}(4),  600--612 (Apr 2004)

\bibitem{xie2022neural}
Xie, Y., Takikawa, T., Saito, S., Litany, O., Yan, S., Khan, N., Tombari, F.,
  Tompkin, J., Sitzmann, V., Sridhar, S.: {Neural fields in visual computing
  and beyond}. Computer Graphics Forum  \textbf{41}(2),  641--676 (May 2022)

\bibitem{zhang2018perceptual}
Zhang, R., Isola, P., Efros, A.A., Shechtman, E., Wang, O.: {The Unreasonable
  Effectiveness of Deep Features as a Perceptual Metric}. In: CVPR (2018)

\end{thebibliography}

\end{document}